# Impact of neural network design principles on the presence of brain-like emergent properties in deep networks for vision

**Niranjan Rajesh, Georgin Jacob & SP Arun**

Centre for Neuroscience, Indian Institute of Science, Bangalore 560012

[#]Correspondence to: sparun@iisc.ac.in

## ABSTRACT

Despite the rapid pace at which deep networks are improving on standardized vision benchmarks, they are still outperformed by humans on real-world vision tasks. One solution to this problem is to make deep networks more brain-like. Although there are several benchmarks that compare the ability of deep networks to predict brain responses on natural images, they do not capture subtle but important emergent properties present in brains. It is also unclear which design principle – architecture, training data or training regime – would have the greatest impact on these emergent properties.

To investigate these issues, we systematically evaluated over 30 state-of-the-art networks with varying network architectures, training datasets and training regimes for the presence/absence of brain-like properties. Our main findings are as follows. First, network architecture had the strongest impact on brain-like properties compared to dataset and training regime variations. Second, networks varied widely in their alignment to the brain with no single network outperforming all others. Taken together, our results offer a principled and interpretable path toward closing the gap between artificial and human vision.

# INTRODUCTION

The vision community is witness to rapid improvements in deep neural networks (DNNs) for visual tasks. These improvements stem from advances in one or more of three key components of the deep learning paradigm: network architecture, training data, or the training regime (Simonyan and Zisserman, 2015; Dosovitskiy et al., 2021; Deng et al., 2009; He et al., 2020; Caron et al., 2021; Zhou et al., 2018; Cao et al., 2018). However, it has been challenging to translate improvements on standard benchmarks to tangible improvements on real-world vision tasks. This difficulty arises primarily from networks being unable to generalize beyond their training distributions (Engstrom et al., 2017; Hendrycks et al., 2021) and their vulnerability to minimally perturbed training samples (Szegedy et al., 2014; Goodfellow et al., 2015). One solution to this puzzling lack of generalization could be to make them more brain-like (Lake et al., 2016; Yamins and DiCarlo, 2016; Hasson et al., 2020; Pramod and Arun, 2022).

There are at least two fundamental issues with making a vision system similar to the brain. First, we need to have effective measures for how similar a deep network is to the brain. Brain similarity has been evaluated by comparing behavioral outputs in categorization tasks or similarity judgments (Wichmann and Geirhos, 2023; Muttenthaler et al., 2025; Pramod and Arun, 2016; Katti and Arun, 2019), or by comparing directly to neural representations (Schrimpf et al., 2020; Willeke et al., 2022; Gifford et al., 2023; Feather et al., 2025; Kornblith et al., 2019; Kriegeskorte, 2008). Although this approach has been fruitful, the outcomes can be highly dependent on the distance metric used (Soni et al., 2024), and can be influenced by stimulus properties, network capacity and even optimization metrics (Pasquiou et al., 2022; Feghhi et al., 2024; Schaeffer et al., 2024).

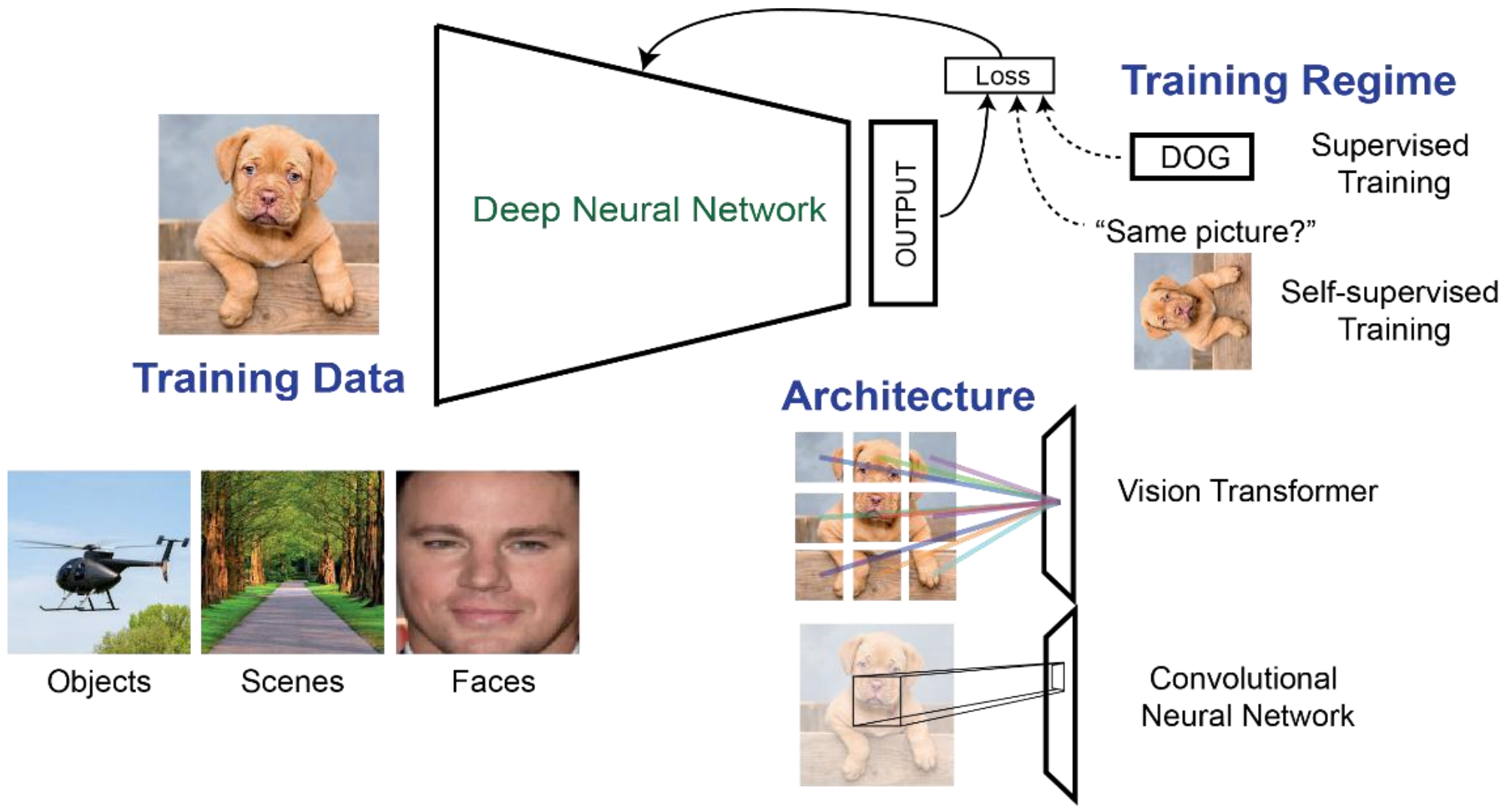

***Figure 1. Key design principles in today's deep networks.*** Deep networks today vary across three key design principles: the overall architecture (*centre*, showing two architectures – convolutional networks and vision transformers), training dataset (*left*, showing three training datasets: objects, scenes and faces) and training regime (*top right*, showing two regimes: self-supervised training and supervised training). Although each design principle is known to affect both overall performance and brain similarity, it is unclear which design principle have the largest impact.

Second, while it makes sense to target changes where they will have the greatest impact, it has been unclear *a priori* which design principle (network architecture, datasets, training) will have the greatest impact on making them more brain-like (Figure 1). Today's deep networks for vision comprise two broad types of architectures with different inductive biases: Convolutional Neural Networks (CNNs) and Vision Transformers (ViTs) (Simonyan and Zisserman, 2015; Dosovitskiy et al., 2021), which have underlying operations (convolution vs attention) that result in divergent internal representations (Raghu et al., 2022). ViTs are argued to be more brain-like in a behavioural sense, showing better robustness to domain shifts and increased sensitivity to object shape relative to texture (Naseer et al., 2021; Tuli et al., 2021). Similarly, aspects of the training regime such the degree of supervision and optimization objectives can affect both the overall performance as well as brain similarity (Khaligh-Razavi and Kriegeskorte, 2014; Wang et al., 2023; Yerxa et al., 2024; Xie et al., 2025). Finally, training data can affect performance as well as brain similarity: training on large diverse datasets with more images commonly encountered by humans can enhance both these desirable features (Mehrer et al., 2020; Muttenthaler et al., 2025), and likewise face-trained networks match face representations in the brain better than object-trained networks (Dobs et al., 2022, 2023).

From the above studies, it is evident that changing network architecture, training dataset and training regime can impact both performance and brain similarity of a deep network. However, these studies only vary one of these design principles while making arbitrary choices for the others. As a result, there are very few comprehensive investigations of which design principle has a larger impact. A recent study has compared a large number of networks for their out-of-distribution performance and robustness to adversarial attacks, showing that the learning task and architecture drives these metrics (Liu et al., 2023). Another recent study showed that, across these design principles, changing the training dataset has the largest influence on brain predictivity (Conwell et al., 2024). Although an insightful step into this direction, the study's analyses are driven by quantitative neural similarity measures which suffer from the limitations outlined above and also do not offer interpretable insight into where these networks diverge from the brain's computations.

**Overview of this study**

As summarized above, to achieve brain-like performance in deep networks, we need a novel approach to evaluate brain similarity and also need to understand which design principles in deep networks will have the greatest impact on brain similarity. We address these two issues in this study.

First, while it is important to evaluate the quantitative similarity between brains and deep networks, we propose that it will also be insightful to evaluate them for the presence or absence of *qualitative* phenomena. For instance, humans show mirror confusion – they confuse mirror images reflected about the vertical axis compared to the horizontal axis (Figure 2). A vision system or network that shows greater vertical axis mirror confusion but perhaps not as strong as humans, would be qualitatively similar to humans and might only require incremental changes to achieve human-like performance. By contrast, a vision system that has the opposite effect (i.e. greater confusion for horizontal mirror images) would likely need more substantial changes to reach human-like performance. Motivated by this idea, we evaluated deep networks on a number of perceptual and neural properties to assess their brain similarity. We have previously reported a number of qualitative similarities and differences between brains and deep networks (Jacob et al., 2021a) and we extend this approach here substantially.

Second, to identify which design principles have the greatest impact on brain similarity, we evaluated many state-of-the-art DNNs varying in their architecture, training regime and training datasets. Our main finding is that network architecture plays the most critical role in driving the emergence of brain-like visual properties. Taken together, our results offer a principled and interpretable path toward closing the gap between artificial and human vision.

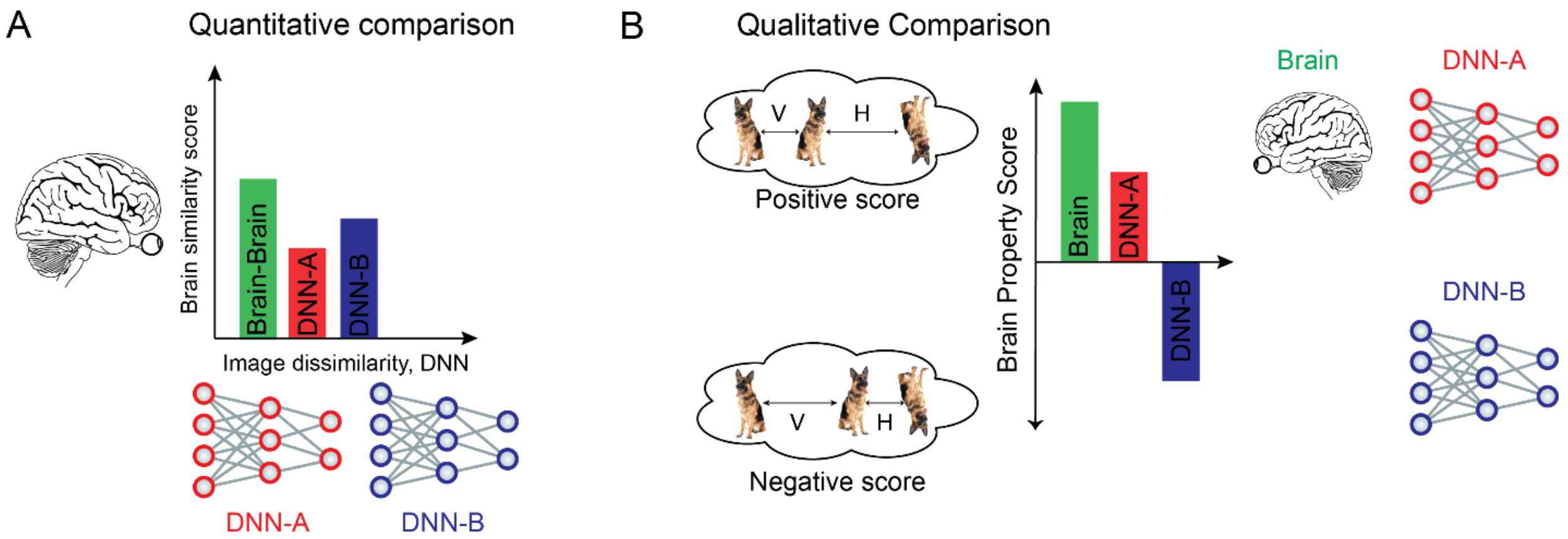

**Figure 2. Quantitative vs qualitative comparisons between brains and networks**

(A) The standard approach in most studies is to compute a brain similarity score by quantitatively comparing natural image representations in brains with deep networks. Thus, DNN-A with a high brain similarity score would be considered more brain-like than DNN-B, with a low score. This resulting brain similarity score can be used to design or find deep networks that are more brain-like.

(B) In this study, we propose that it could be also insightful to evaluate deep networks for qualitative properties that are known to be present in brains. For example, it is well known that vertical mirror images are more similar compared to horizontal mirror images. Such an effect can be quantified by a *Brain Property Score*, which is positive if brain-like and negative if the effect is opposite. Such a qualitative comparison could lead to very different conclusions. For instance, DNN-A, might have the same mirror confusion but the effect might be weaker but in the same direction as the brain, and is therefore brain-like – and might be improved by incremental changes. By contrast, DNN-B, which has a negative score, has the opposite effect and might require more drastic changes to make it more brain-like.

## RESULTS

We tested a total of 32 deep networks for vision with variations in network architecture, training dataset and training regime for the presence or absence of a large number of emergent brain-like properties. In each case, we evaluated deep networks varying in each design principle while holding other choices constant to study the impact of that factor on all the brain properties tested. Table 1 summarizes each property briefly (for more details, see Methods).

| Property Name | Brief description |
|---|---|
| Object Normalization-Pairs | Units response to two objects is the average of the individual object responses |
| Object Normalization-Triplets | Units response to three objects is the average of the individual object responses |
| Scene Incongruence | Object recognition accuracy drops for objects presented against incongruent compared to congruent backgrounds. |
| Mirror Confusion | Distances between vertical mirror images is smaller than distances between horizontal mirror images. |
| Morphline Sparseness | Units responding sparsely to morphed shapes also do so for other morphed shapes, showing correlated sparseness. |
| Shape/Texture Sparseness | Units responding sparsely to shapes also do so for textures, showing correlated sparseness. |
| Weber's Law | Distances between objects differing in line length depends on relative, not absolute, changes in magnitude. |
| Basic Occlusion Processing | Distance between an occluded display and the revealed objects is smaller than equivalent image changes. |
| Depth Occlusion Processing | Distances between occluded displays that differ in depth ordering is smaller than equivalent image changes. |
| Relative Size Encoding | Units are sensitive to the relative size of features rather than absolute size. |
| Surface Invariance | Units respond to congruent surface and pattern changes |
| 3D Processing 1 | Units show greater sensitivity to 3D shape changes than to 2D changes. |
| 3D Processing 2 | Distances between features that change 3D shape are larger than equivalent 2D shapes, even with clutter controlled. |
| Global Processing | Distances between shapes differing at the global level are larger than distances between shapes differing at the local level. |
| Thatcher Faces | Distances between upright faces is larger than distances between inverted faces. |

**Table 1. Summary of brain-like emergent properties tested in deep networks.**

For each property, we captured unit activations from the penultimate densely connected layer for a specific stimulus set and computed the effect strength. The same property value for the brain, reported from previous studies, was then used to normalize each of the models' effect strengths to compute a Brain Property Score such that the brain itself would have a score of 1. A detailed account of the brain properties chosen and how the effect strengths were computed is outlined in the Methods section and Supplementary Section S1. An additional property of Scene Incongruence was tested; since it involved comparisons of classification accuracies of objects, it was limited to the architecture manipulation only where the classifier output remained unchanged.

With a Brain Property Score for each network, we were able to compare the model's performance in that tested property with other networks as well as the brain. The effect strength for every property was always normalized to remove any dependence on the absolute magnitude of activations in networks or the brain. For ease of interpretability, the scores were computed such that positive values are always brain-like, and negative values are anti-brain-like.

**Impact of Network Architecture on Brain-Like Properties**

We first examined how the inductive biases inherent in different architectural designs influence the emergence of brain properties in DNNs. We varied the fundamental visual operation underlying modern vision architectures by selecting DNNs from two distinct families: CNNs, which rely on convolutional operations for feature extraction (Krizhevsky et al., 2017) and ViTs, which utilize self-attention mechanisms to process visual information (Dosovitskiy et al., 2021)

Our analysis included networks of varying sizes from state-of-the-art models in both families: VGGs (Simonyan and Zisserman, 2015), ResNets (He et al., 2016) Inception Nets (Szegedy et al., 2015, 2016), ConvNeXts (Liu et al., 2022), Vanilla ViTs (Dosovitskiy et al., 2021), SWIN (Liu et al., 2021), and DeiT (Touvron et al., 2021). All networks were trained on ImageNet-1K (Deng et al., 2009) in a supervised manner, ensuring that only architectural differences, drive any observed differences in brain similarity.

Comparing Brain Property Scores across networks revealed several interesting patterns (Figure 3), which we detail below.

*Universally Present Properties Across Architectures.* Both CNN and ViT families consistently exhibited certain brain-like properties, including Object Normalization, Scene Incongruence, Mirror Confusion, and Correlated Sparseness. The universal emergence of these properties suggests they arise naturally from optimizing for object recognition regardless of training architecture. Object Normalization likely emerges because both architectural families' individual units develop approximately linear superposition properties in their learned representations (Olah et al., 2020) where multi-object scenes are processed as combinations of individual object responses. This linearity may arise naturally from the distributed nature of both convolutional and attention-based feature representations in earlier layers. Scene Incongruence effects reflect both architectures' ability to learn statistical regularities between objects and their typical contexts through ImageNet training, where contextual priors improve classification accuracy. Mirror Confusion patterns suggest that both architectures develop similar orientation-sensitive features and a preference for vertical axis reflections perhaps due to the inclusion of standard augmentations during training. The higher-than-brain property scores for Mirror Confusion could be an indication of an over-reliance of vertical reflections during training.
*Universally Absent Properties.* Conversely, several brain-like properties were consistently absent across both architectural families. Properties including relative size encoding, surface invariance, 3D processing, global shape processing, and Thatcher face processing were missing from nearly all tested networks. These absences reveal fundamental differences between current deep networks and brains.

The absence of these properties may reflect fundamental limitations in current training paradigms and architectural designs. 3D processing deficits likely stem from ImageNet's 2D nature, where depth information is implicit rather than explicit, preventing networks from developing robust 3D shape sensitivity. Surface invariance and relative size encoding may require understanding of shape and scene properties that are not explicitly rewarded by 2D image classification tasks. The lack of global shape processing in most networks suggests that local feature extraction (CNNs) and patch-based processing (most ViTs) may interfere with developing holistic shape representations. Thatcher face processing deficits almost certainly stem from the lack of face data in the DNNs' visual diet which is required to learn the facial features relevant to the property.

*Architecture Differences in Brain Similarities.* Despite these commonalities, the two architectural families showed distinct patterns of brain similarity. CNNs consistently exhibited closer alignment to brain responses in correlated sparseness experiments compared to ViTs. This difference likely stems from CNNs' hierarchical feature learning.

Since ViTs pool and process global information, they tend to also maintain distributed feature representations (to a greater degree than in CNNs) – meaning that single units do not necessarily become highly selective for singular features which could explain why there is less correlation in selectivity as observed in the sparseness properties (Raghu et al., 2022).

CNNs also showed better adherence to Weber's Law compared to ViTs, which consistently failed to exhibit this psychophysical principle. Convolutional processing of image information leads CNNs to synthesize its representations of a full image incrementally, unlike in ViTs, which could also make it more sensitive to relative differences in feature sizes. Interestingly, only vanilla ViT models approached the brain's global shape processing advantage. This finding aligns with ViTs' global attention mechanism, which can simultaneously integrate information across all spatial locations from the earliest layers, potentially enabling better holistic shape processing compared to CNNs' locally constrained receptive fields that must build global representations through multiple hierarchical stages. This advantage disappears for later ViTs like Swin and DeiT perhaps due to their convergence to CNN-like inductive biases as mentioned below.

*Visualization of similarities between networks and brains.* To visualize the relative similarities between the effect strengths of the brain and all the tested deep networks, we performed a principal component analysis (PCA) and embedded them along the first two dimensions. This revealed strong clustering by architectural family (Figure 3B), with two notable exceptions: vanilla ViTs and ConvNeXts. Vanilla ViTs performed remarkably better than other ViT variants in Mirror Confusion and Global Processing while maintaining similar performance on other properties. This superior performance may reflect the vanilla ViT's pure attention mechanism, unencumbered by the CNN-like inductive biases introduced in Swin (hierarchical processing with sliding windows) and DeiT (distillation from CNN representations).

ConvNeXts clustered closer to ViTs than to their CNN relatives, consistent with their design philosophy of "modernizing" CNNs by incorporating ViT-inspired design principles (Liu et al., 2022). The architectural modifications in ConvNeXt—including larger kernel sizes, fewer activation functions, and LayerNorm—may have shifted their representational properties toward those of attention-based models.

The intermediate positioning of SWIN and DeiT architectures between pure CNNs and vanilla ViTs further supports the hypothesis that architectural modifications systematically influence brain-like properties. SWIN's hierarchical windowed attention reintroduces CNN-like local processing biases, while DeiT's distillation training explicitly encourages CNN-like representations.

*Conclusion*

We conclude that network architecture causes a large variation in the presence or absence of brain-like emergent properties, with some properties universally present, others universally absent and others that are present in some architectures but not others.

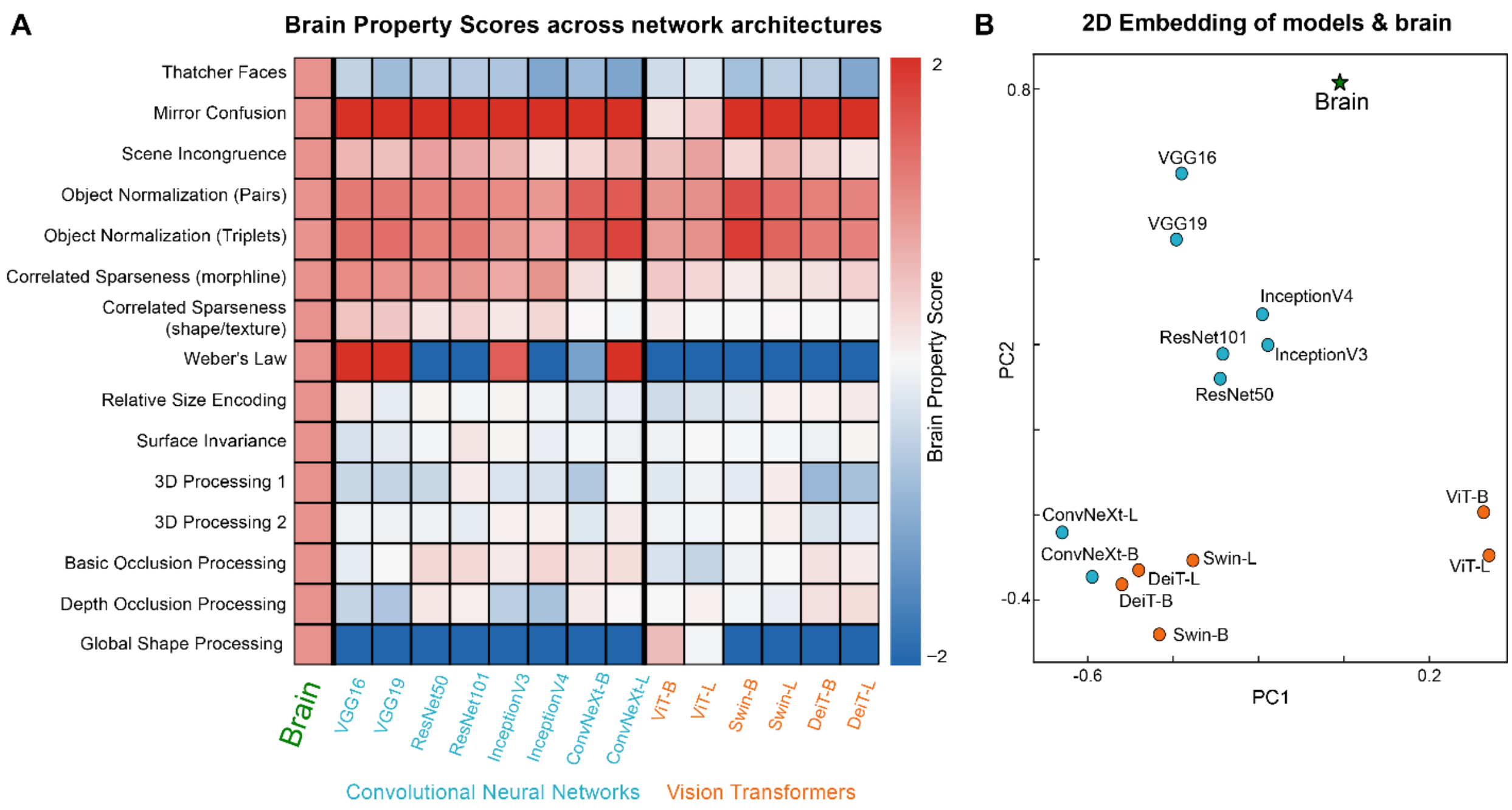

**Figure 3 Influence of network architecture on brain-like properties.**

(A) Brain property scores for 14 networks from two broad deep network architectures (CNN and ViT). The color of each cell indicates the brain property score for each property and network limited to the range of -2 and 2. All brain property scores were computed by normalizing model effect strengths with the empirically measured scores for the brain obtained from previous studies.

(B) To visualize the similarity of these networks to each other and to the brain, we projected the brain property scores of each vision system along their first two principal components (PC1 & PC2). Nearby points indicate networks with similar Brain Property Scores across all tested properties. Thus, CNN & ViT architectures form separate groups, suggesting that they exhibit similar scores, but neither group is systematically closer the brain along all the tested properties.

**Impact of Training Dataset Variations on Brain-Like Properties**

Next, we investigated the effect of visual experience during training in the emergence of brain properties, we selected two CNNs and a ViT, each trained on ImageNet-1K for objects, Places365 for scenes, and CASIA & VGGFaces2 for faces (Muhammad et al., 2021; Cao et al., 2018; Zhong and Deng, 2021; Schroff et al., 2015). We evaluated these 9 networks for the presence/absence of brain-like properties as before. The results are summarized in Figure 4.

*Universal Properties Across Training Datasets.* Consistent with our architectural findings, Object Normalization (both pairs and triplets), Mirror Confusion, and Correlated Sparseness for morphlines emerged universally across all training datasets (Figure 4A). However, a notable exception emerged: face-trained networks failed to exhibit correlated selectivity for shapes and textures. This absence likely reflects the constrained feature space of face datasets, where the statistical correlation between shape and texture features differs fundamentally from natural object categories. Face recognition may rely on more specialized feature combinations (e.g., spatial relationships between facial landmarks) rather than the general shape-texture correlations that emerge from diverse object recognition.

*Universally Absent Properties Across Training Datasets.* Relative Size Encoding, Surface Invariance and 3D processing are once again absent across all datasets. These persistent absences across all training suggest that current 2D image-based training paradigms -- regardless of content domain or architecture -- cannot provide the learning signals necessary for these particular brain-like computations.

*Training Dataset Differences in Brain Similarity.* When varying network architectures, we noted that none of the Vision Transformers exhibited Weber's law (Figure 3A). However, we now note that training on scenes and faces causes the emergence of Weber's law in ViTs (Figure 3A). In fact, the scene-trained ViT is closest to the true brain-score for Weber's Law. Scene datasets may provide richer scale variation and spatial context that encourages proportional size encoding, while face datasets contain systematic variations in facial feature proportions that could drive similar sensitivity. On the other hand, face and scene-trained CNNs lost the ability to process occlusion when compared to object training. Scene and face datasets may lack this diversity of occlusion patterns. The scenes typically show complete objects in context, while face datasets focus on frontal, unoccluded presentations. 3D Processing showed a slight and selective emergence only in scene and face-trained networks. Scene datasets inherently contain more depth cues and spatial relationships that could encourage 3D-sensitive representations, while face datasets' lack of feature diversity may encourage networks to also pick up depth relationships that are not explicitly conveyed in the data. Finally, and as expected, the Thatcher Effect emerged exclusively in face-trained networks, confirming our hypothesis that domain-specific visual experience is necessary for specialized processing mechanisms. Face datasets provide the statistical regularities in facial feature configurations required to develop sensitivity to human-like face-specific geometric distortions.

*Visualization of similarities between networks and brains.*

To visualize the similarities between brains and deep networks, we projected the vector of brain property scores across networks along the first two principal components. This revealed weak clustering by training dataset compared to the strong architectural clustering observed previously (Figure 3B). However, face-trained networks showed distinct separation from object and scene-trained networks within each architectural family. This pattern suggests that face processing develops qualitatively different representational structures compared to general object or scene recognition, possibly reflecting the specialized computational demands of face recognition that require encoding of subtle metric relationships rather than categorical distinctions.

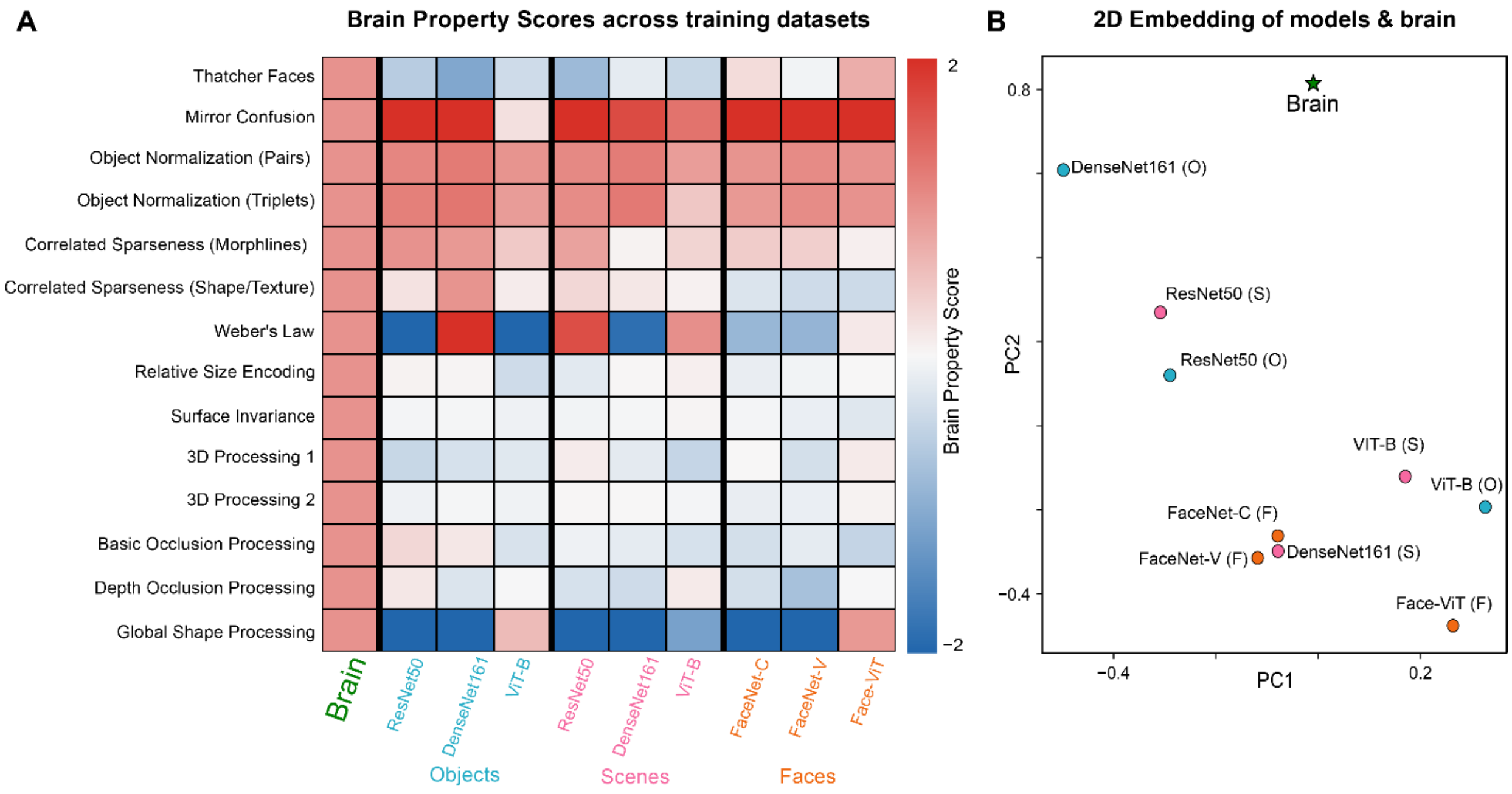

**Figure 4 Influence of training data on brain-like properties**

(A) Colormap of Brain Property Scores for 9 deep networks trained on different datasets: ImageNet-1k (Objects), Places365 (Scenes) and CASIA/VGG-Face2 (Faces) on each brain property.

(B) Visualization of all tested networks along their first two principal components. Nearby points indicate networks with similar Brain Property Scores across all tested properties.

*Conclusions*

We conclude that training datasets causes a relatively smaller variation in the presence or absence of brain-like emergent properties compared to changing network architecture. Here too, some properties were universally present, some were universally absent, and others were present in some datasets but not others.

**Impact of Training Regime Variation on Brain-Like Properties**

Finally, we investigated the effect of training regime in the emergence of brain properties. We chose a ResNet50 and a vanilla ViT-base network across different training regimes on the same dataset - ImageNet1k. In the supervised category, we test the standard trained networks and their adversarially trained counterparts. Adversarial training involves injecting adversarial noise during training to prevent learning of non-robust features as a defense against adversarial examples (Madry et al., 2018; Ilyas, 2019). In the self-supervised category, we tested both architectures using two popular self-supervised learning algorithms: MOCOv3 and DINO (Caron et al., 2021; Chen et al., 2021). Both self-supervised algorithms encourage networks to learn similar representations for different views of the same image without explicit labels. We evaluated these 8 networks for the presence or absence of brain-like properties as before.

*Universal Properties Across Training Regimes.* Consistent with the previous investigations, Object Normalization (both pairs and triplets), Mirror Confusion, and Correlated Sparseness (both morphlines and shape/texture variants) emerged universally across all training regimes (Figure 5A). This reinforces our earlier hypothesis that these properties arise naturally from optimization pressure for visual recognition tasks, independent of specific architectural mechanisms or training objectives.

*Universally Absent Properties Across Training Regimes.* Surface Invariance, 3D Processing (both variants), and Global Shape Processing remained consistently absent across all training paradigms, confirming the limitations identified in our architectural analysis.

*Training Regime Differences in Brain Similarity.* Self-supervised networks (MOCOv3 and DINO) exhibited slightly stronger adherence to Weber's Law compared to supervised counterparts. Contrastive learning's emphasis on scale invariance during data augmentation may encourage networks to encode proportional rather than absolute feature differences, as the same object must be recognized across multiple scales within the same training batch. Conversely, supervised networks showed superior performance in both Basic and Depth Occlusion processing. Explicit object categorization objectives may drive networks to develop more robust mechanisms for inferring occluded object completions, as accurate classification often requires understanding whole objects despite partial visibility—a requirement less critical for contrastive learning objectives.

Interestingly, the adversarially-trained ViT and the MoCo-trained ViT loses the global shape processing advantage possessed by the other vanilla ViTs. This could be attributed to pressure to learn localized features during contrastive and adversarial learning – something which may be corrected in the distillation process of the DINO-trained ViT.

*Visualization of similarities between deep networks and brains*

Overall, we saw fewer variations in qualitative effects on varying the training regime. This is further illustrated by the embedding plot in Figure 3B where the models cluster based on architecture rather than training regime. In the case of ResNets, we do see the MoCo and DINO variants closer together and the same for their supervised counterparts, but this is absent for Vision Transformers.

*Conclusions*

We conclude that training regime causes a relatively smaller variation in the presence or absence of brain-like emergent properties compared to changing network architecture or training dataset. Here too, some properties were universally present, some were universally absent, and others were present in some architectures but not others.

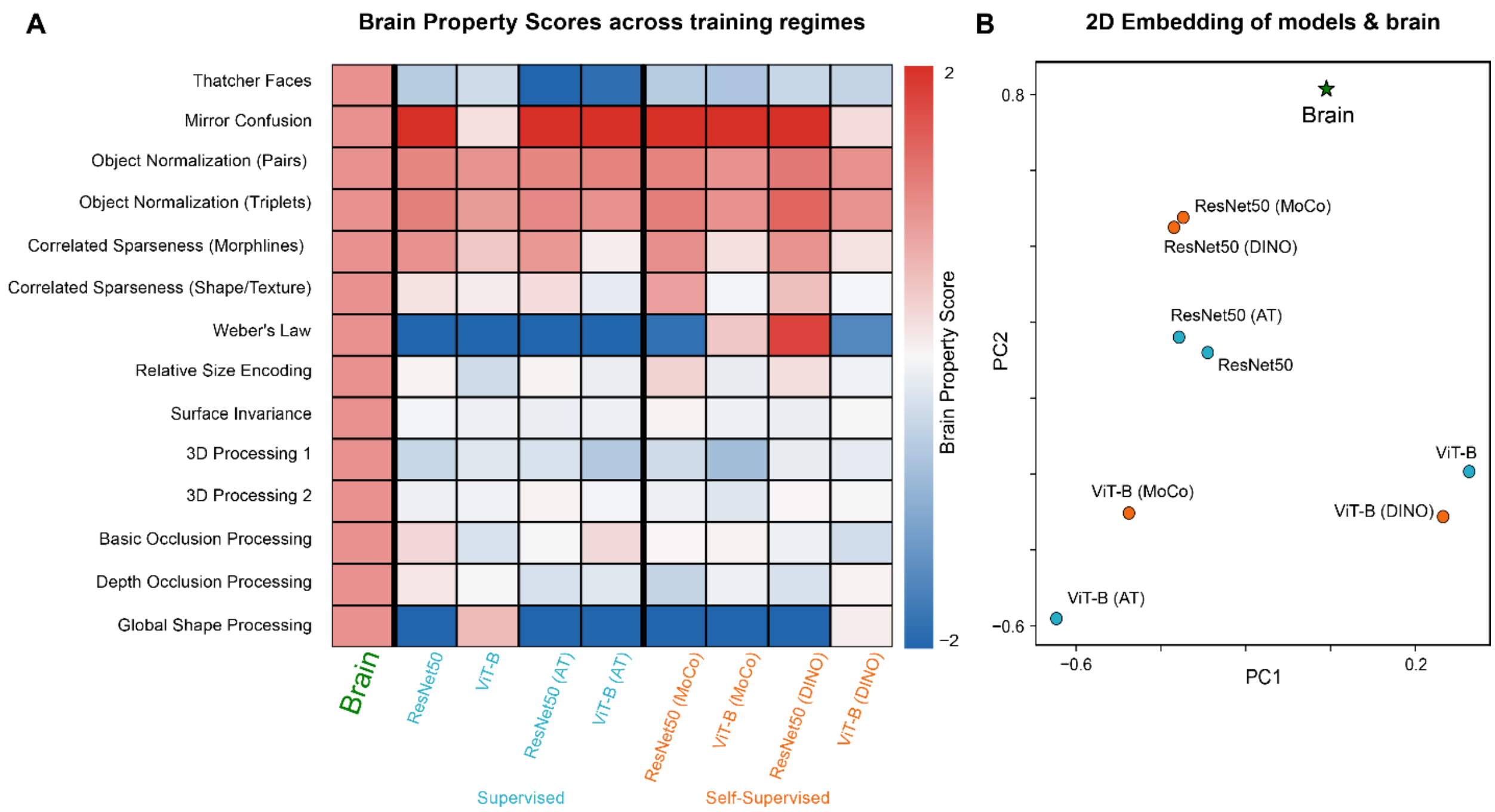

**Figure 5. Influence of training regime on brain-like properties**

(A) Colormap of Brain Property Scores for 8 networks trained using different algorithms: standard and adversarial training for the supervised group, MoCov2 and DINO for the self-supervised group on each brain property.

(B) Visualization of all tested networks along their first two principal components. Nearby points indicate networks with similar Brain Property Scores across all tested properties.

**Brain-like properties across all networks**

The results of the preceding sections are based on careful comparison of varying each design principle while holding other factors constant. To perform a more global comparison across networks, we created an embedding of all the tested networks along their principal component dimensions (see Section S1 for a global summary of all brain-like properties across all networks). In addition, we included popular brain-aligned networks such as CORNet (Kubilius et al., 2019) and VOneNet variations (Dapello et al., 2020). This allows us to compare whether optimizing for quantitative brain similarity leads to a greater prevalence of brain-like properties.

As noted previously in the individual variation results, the biggest driver of variation in the Brain Property Score space is architecture, which is evident in Figure 6A (blue and orange markers). The grouping based on other design principles is weaker. The brain-aligned networks also seem to be qualitatively similar to their standard counterparts: CORnet-RT (with residual connections inspired by ResNets) and VOneResNet50 position themselves close to other DNNs we tested from ResNet family. This suggests that current brain-aligned networks do not fundamentally alter the qualitative brain properties we measured. This finding also reinforces our observation of network architecture being the strongest influence on the emergence of Brain Properties in DNNs.

To quantify our observation that architecture has the largest effect, we calculated the between-group distance for deep networks organized by each design principle. This revealed significantly larger between-group distances for deep networks organized by architecture compared to those organized by training dataset ($p < 0.005$, rank-sum test) and compared to those organized by training regime ($p < 0.005$, rank-sum test).

We also asked which design principle had networks that were most similar to the brain. To this end, we calculated the cosine similarity between each network's effect strength vectors (made up of a model's effect strength for all tested properties) and the brain (Figure 6C). We see that in architecture, CNNs are more brain-like than ViTs; in training datasets, object and scenes datasets outperform the face-trained networks; and in training regimes, networks in both groups exhibit similar levels of brain-likeness. We also note that within each design principle, architecture seems to have the most diversity in terms of brain-similarity given that we exclude face-trained networks from the training dataset class due to very low property similarity with the brain. This finding suggests that architecture does not only drive the emergence of brain properties, but also that architectural addition of convolutional filters lead to properties most similar to the true brain.

**Relation between brain properties and generalization/robustness**

Next we wondered whether the brain property scores of deep networks are related to other relevant metrics of their performance, in particular their ability to generalize to out-of-distribution (OOD) images, or their vulnerability to adversarial examples (Goodfellow et al., 2015; Kar et al., 2022). Across the tested networks for which these metrics were available, we observed that some brain property scores were significantly correlated with these metrics (Section S3).

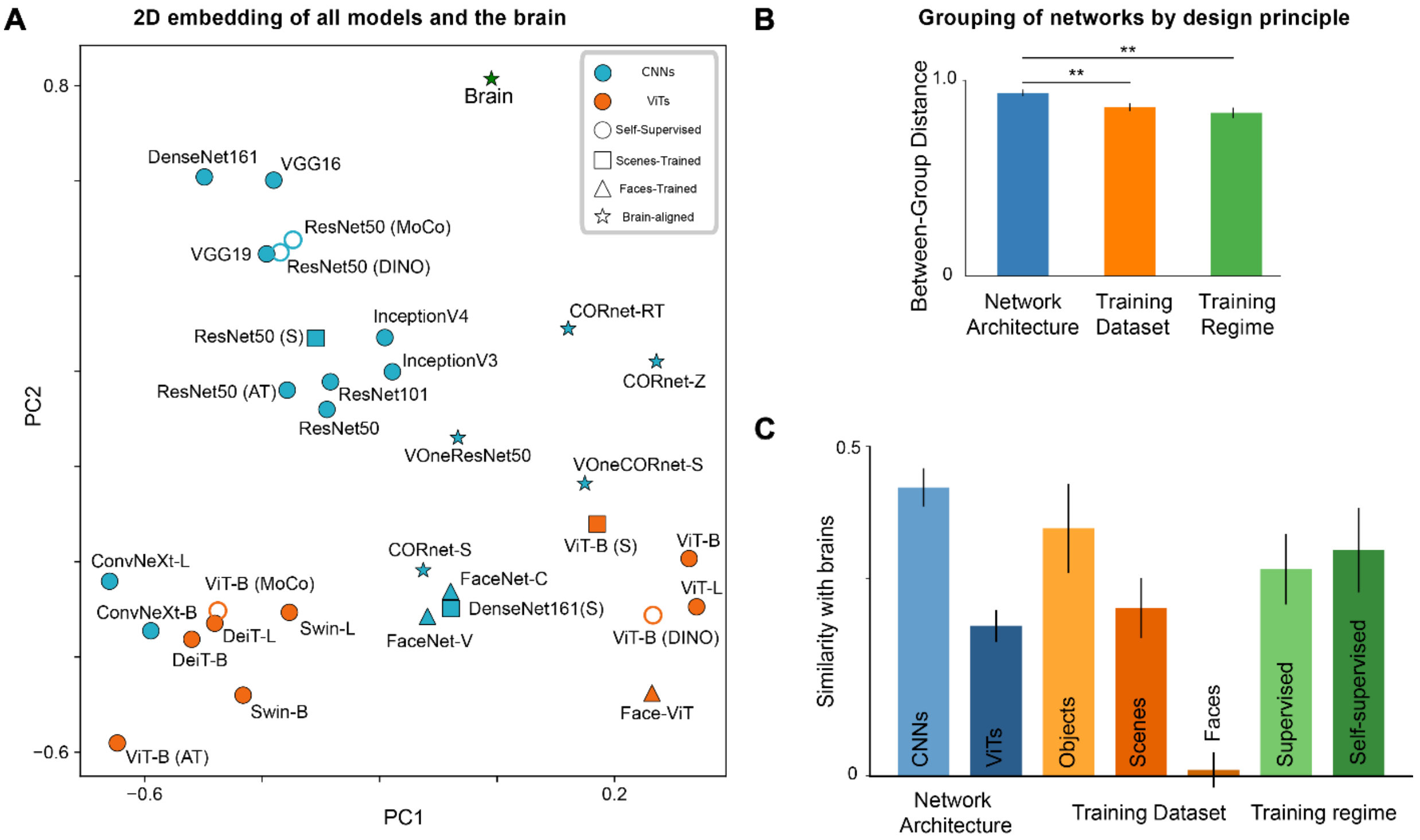

**Figure 6. Similarity between all tested networks on brain-like emergent properties**

(A) Visualization of all tested networks along their first two principal components. Nearby points indicate networks with similar Brain Property Scores across all tested properties.

(B) Mean between-group distance between networks sharing a particular design principle, showing greater clustering of networks by network architecture, and then by training dataset, and relatively little by training regime.

(C) Mean similarity between Brain Property Score vectors of deep networks in each group with the brain.

## GENERAL DISCUSSION

In this work, we describe a set of brain-like emergent properties that can be evaluated on any deep network, or for that matter, any visual system with an accessible internal representation. We systematically tested these properties on 32 state-of-the-art DNNs that varied in their architecture, training dataset and training regime. Our main findings are as follows. First, network architecture had the strongest impact on the presence or absence of brain-like properties compared to dataset and training regime variations. Second, networks varied widely in their alignment to the brain with no single network achieving a very close match.

Our study offers interesting insights into the design choices under which the brain-like properties arise. Properties like object normalization and mirror confusion are present in all networks, suggesting that they emerge with the demand of image recognition. Some properties such as relative size, surface invariance and 3D processing are almost always absent, suggesting that they may emerge only with specialized training. Some properties like Weber's law are present only in convolutional networks, whereas global advantage is present in vanilla ViTs, suggesting that these effects arise largely due to network architecture. Finally, training dataset also matters, since face-trained CNNs lose occlusion processing and shape-texture sparseness while gaining the Thatcher effect. We also see interesting interaction effects like in the case of Weber's law where object-trained vanilla ViTs do not exhibit the property but after training with scenes and faces data, the property starts emerging (Figure 4A).

Taken together, our results offer insights into the presence of brain-like emergent properties in deep networks. They raise the intriguing possibility that training deep network to acquire these properties could lead to more generalizable and robust brain-like deep networks.

## METHODS

### Brain Properties

We adapted 15 well-known properties from visual psychology and neuroscience from the work of (Jacob et al., 2021a). A DNN could be scored on each brain property in the range of [-1, +1]. Positive effect strengths indicate a presence of the property, making the network brain-like. Negative scores indicate that the network/model has an opposite property i.e. is anti-brain like. We also report an empirically observed score for each of these property from human behavior or monkey high-level visual areas to facilitate a direct comparison to the brain. Model effect strengths are normalized by the observed brain score for each property to obtain the Brain Property Score. A summary of each of the brain properties for a model is given in Table 1.

These experiments were adapted from a recent study comparing deep networks with emergent brain-like properties (Jacob et al., 2021b). To test each property on DNNs, we conducted an experiment with a specialized stimulus set used for analogous behavioral or neurophysiological studies. For the scene incongruence experiment (BP3), the DNN's categorization performance on the scene incongruence sets was used as input to compute the effect strength. For the rest of the properties, the DNN's penultimate layer's activations to their corresponding stimulus sets were used as input. We now outline the methodology for computing the effect strength of each property.

### Thatcher Faces (BP1)

Another popular qualitative effect in human visual processing is that we are more sensitive to facial feature changes on an upright face than an inverted face (Bartlett and Searcy, 1993). We measure this effect in DNNs by computing a Thatcher Effect ($P_{thatcher}$) by considering the difference between an upright face and the same face with flipped features as $d_u$ and the difference between an inverted face and the inverted face with flipped features $d_i$.

$$P_{thatcher} = \frac{d_u - d_i}{d_u + d_i}$$

**Mirror Confusion (BP2)**

In this experiment, we verify whether the neural representations of a DNN for a vertically flipped image is closer to the original image than a horizontally flipped image. This effect is commonly observed in human behavior and monkey neural data and is termed the mirror confusion effect (Rollenhagen and Olson, 2000). We compute a mirror confusion index ($M$) as the following modulation index where $d_h$ and $d_v$ are the distances between the horizontally flipped image and the vertically flipped image from the original image respectively.

$$M = \frac{d_h - d_v}{d_h + d_v}$$

**Scene Incongruence (BP3)**

A famous observation is the effect context plays in visual categorization tasks in humans. When the background of an image is incongruent, in the semantic sense, to the object in the foreground, humans see a fall in their categorization performance (Davenport and Potter, 2004; Munneke et al., 2013). We test the same effect here and compute a scene incongruence index ($P_{scene}$) with $a_c$ being the average accuracy of the network on congruent object-scene images and $a_i$ being the incongruent counterpart and taking the ratio of their differences:

$$P_{scene} = \frac{a_c - a_i}{a_c + a_i}$$

As scene incongruence was only comparisons of classification accuracies of objects, it was only included in the architectural comparison.

**Object Normalization (BP4 – Pairs, BP5 – Triplets)**

In this experiment, we verify whether the neural response for a group of objects is an average of the individual object responses -- an effect observed in high-level monkey visual regions (Zoccolan et al., 2005). To measure this effect, we identify visually active neural units for all possible positions within the stimulus where an object may appear and record the individual object responses and the responses when all objects are shown together in the same stimulus. Each stimulus had three positions along the center. For the single stimuli, an image was shown at only one of these positions. For the pairs, two images were shown at two of these positions. For triplets, three images were shown at three of these positions. We then compute the average slope of the multiple object response plotted as a function of the sum of individual object responses. If the network exhibits multiple object normalization, we should observe an average slope of 0.5 for pairs of objects and 0.33 for object triplets.

**Correlated Sparseness (BP6 – Morphlines, BP7 – Shape/Texture)**

For the Morphlines experiment (BP6), visually active units for certain shapes are identified and are shown parametric variations of the same shape (morphlines). To show the brain-like effect of correlated morphline sparseness, these units that are selective for (or exhibit sparse responses) for only a few shapes should also be highly responsive to parametric changes of that shape (Zhivago and Arun, 2016). In our experiment, we show images from a reference set to the DNN and a parametric morphline of the same images. Neural unit sparseness on the reference set is computed and correlated with the sparseness on the parametrically varying set to arrive at a correlated sparseness strength.

For the Shape/Texture experiment (BP7), units in the brain that are shown to respond sparsely to shapes (selective to shapes) are also sparsely responsive to textures (Zhivago and Arun, 2016). The sparseness values for visually active units for shapes and stimuli are computed and correlated to get the correlated sparseness effect strength.

We report the correlation of sparseness values across units for the corresponding stimuli sets as the brain property scores.

**Weber's Law (BP8)**

To measure the effect strength of this fundamental property in human visual behavior, we show images of varying stimuli length to the network and correlate the representational difference between the stimuli and the corresponding absolute and relative length changes (Pramod and Arun, 2014). By taking the difference between the relative correlation and absolute correlation, we verify if the network is sensitive to relative length differences or absolute length differences.

**Relative Size Encoding (BP9)**

Neurons from monkey visual areas have been shown to encode relative size (Vighneshvel and Arun, 2015). For example, two parts of an object proportionally varying in size lead to a different magnitude of neural activation changes as compared to disproportional changes in size. To compute the strength of this effect we take $d_1$ to be the neural distance between objects with disproportional change in part sizes and $d_2$ to be the neural distance between objects with proportional changes in part size and take the ratio of differences as the relative size index ($P_{relsize}$):

$$P_{relsize} = \frac{d_2 - d_1}{d_2 + d_1}$$

**Surface Invariance (BP10)**

Similarly, neurons also show that consistent changes between an object and the surface it rests on is closer in the neural space than inconsistent changes between the two (Ratan Murty and Arun, 2017). We compute the surface invariance modulation index ($P_{surf}$) by taking $d_1$ as the neural activation distance between stimuli with inconsistent or incongruent changes in surface and object shape and $d_2$ to be the distance between stimuli with the same surface and shape change:

$$P_{surf} = \frac{d_2 - d_1}{d_2 + d_1}$$

**3D Processing (BP11, BP12)**

3D objects are observed to be more distinct in human perception compared to similar corresponding 2D (Enns and Rensink, 1990, 1991). We test this effect in DNNs by computing the modulation index for 3D processing, as a ratio of differences $d_1$ - the neural activation distance between 3D objects and $d_2$ - the distance between 2D objects of equivalent pixel or feature differences:

$$P_{3D} = \frac{d_1 - d_2}{d_1 + d_2}$$

**Occlusion Processing (BP13 – Basic, BP14 – Depth)**

In human perception, images of objects and the same objects occluding each other are closer together than the equivalent 2D feature distances without occlusion between the two images (Rensink and Enns, 1998). To verify if this effect exists in DNNs, we compute a modulation index below where the Occlusion Index $(P_{occ})$ using the neural activation differences between the objects with and without occlusion $(d_1)$ and the same objects without occlusion and the equivalent 2D feature distances from the occluded image $(d_2)$

$$P_{occ} = \frac{d_1 - d_2}{d_1 + d_2}$$

The same experiment with a different stimuli set of depth-based occlusions is tested for BP14.

**Global Shape Processing (BP15)**

Humans are more sensitive to the global shape of objects than local shapes that make up the same global shape (Jacob and Arun, 2020). In perceptual space, the distance between two configurations with global changes is greater than two configurations with a local change. We compute a global processing index $(P_{global})$ by taking the neural activation differences between stimuli with global $(d_g)$ and local changes$(d_l)$:

$$P_{global} = \frac{d_g - d_l}{d_g + d_l}$$

**Model Selection**

Throughout the study, we selected 32 pre-trained, openly available, deep networks. All Imagenet-trained DNNs were imported from the Pytorch Image Models Repository (TIMM) (Wightman et al., 2021). The remaining DNNs were sourced from their official repositories.

**Model Embeddings**

For each DNN, a Brain Property Score was computed for each of the measured properties. Combining them, a DNN's Brain Property Score vector was obtained for each of the 32 DNNs. Each vector was 14-dimensional (accounting for the 14 Brain Properties that can be tested for all DNNs). We also had a corresponding vector for the brain's true values. Principal Component Analysis (PCA) was employed to project the 33 points onto a 2D plane. These PC coordinates were used for visualization.

**Grouping effect of each design principle**

In order to determine how each design principle contributes to the emergence of all the measured Brain Properties, we conducted a grouping effect analysis. This entailed treating all the 32 DNNs Brain Property Vectors as points on a vector space and assigning corresponding labels for three grouping configurations: Architecture, Dataset and Regime. For each configuration, the mean distance between all pairs of points belonging to different label groups was computed. These ‘between-group’ distances were used to quantify the grouping effect of each design principle.

### Similarity of brain property scores of each network with the brain

We also measured how close each design factor’s brain property scores were to those of the observed, true brain values. For this, we computed the cosine similarities for each model and averaged across the models that belong to each design factor (CNN, ViT for architecture; Objects, Scenes and Faces for dataset; Supervised and Self-Supervised for regime).

## ACKNOWLEDGEMENTS

This research was supported by the DBT/Wellcome Trust India Alliance Senior Fellowship (Grant# IA/S/17/1/503081) and a Google Asia-Pacific grant awarded to SPA.

## DATA AVAILABILITY

All data and code required to reproduce the results are publicly available at https://osf.io/tjqp8/

**SUPPLEMENTARY MATERIAL**

**for**

# Impact of neural network design principles on the presence of brain-like emergent properties in deep networks for vision

**Niranjan Rajesh, Georgin Jacob & SP Arun**

**CONTENTS**

## SECTION S1 – BRAIN PROPERTY SCORES FOR ALL 32 NETWORKS

To round out our analyses, we report the brain property strengths for all the tested networks.

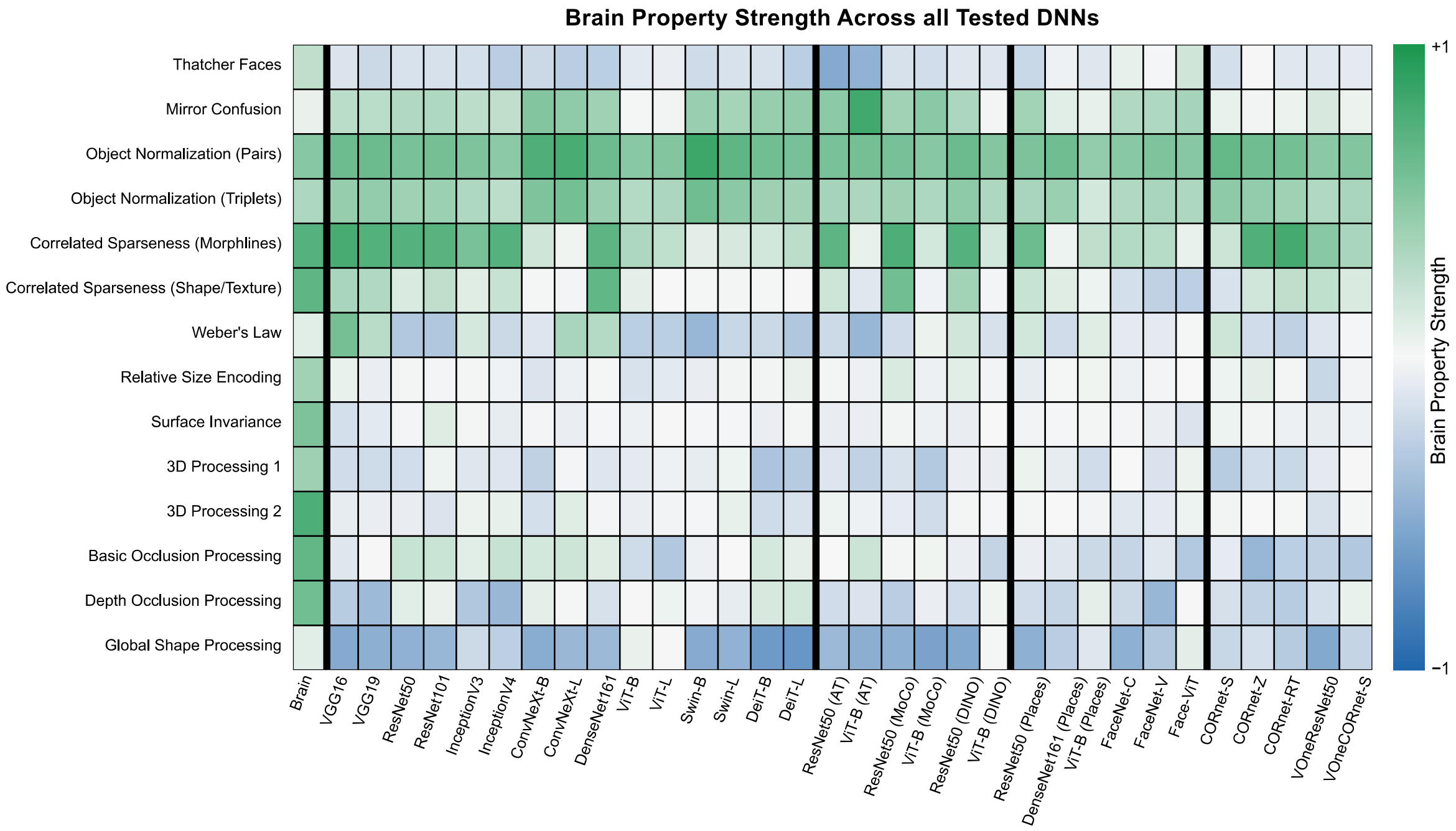

**Figure S1. Brain property strengths for all 32 deep networks tested in this study.** Each cell represents a model's (row) effect strength on the tested property (column). These effect strengths were also empirically computed for the brain from previous studies. Only 14 out of the 15 properties are shown here as Scene Incongruence was only measured for models in the architecture variation investigation.

## SECTION S2. LAYERWISE EVOLUTION OF BRAIN PROPERTIES

To visualize how the brain properties evolve across the layers of two example networks, we tested a VGG16 and a vanilla ViT-base on all the 14 brain properties (we excluded scene incongruence as it does not have a layer-wise effect). For most properties, brain property strengths reach positive (brain-like) levels only in the final layers, underscoring the fact that they are emergent properties and not a trivial consequence of early visual processing.

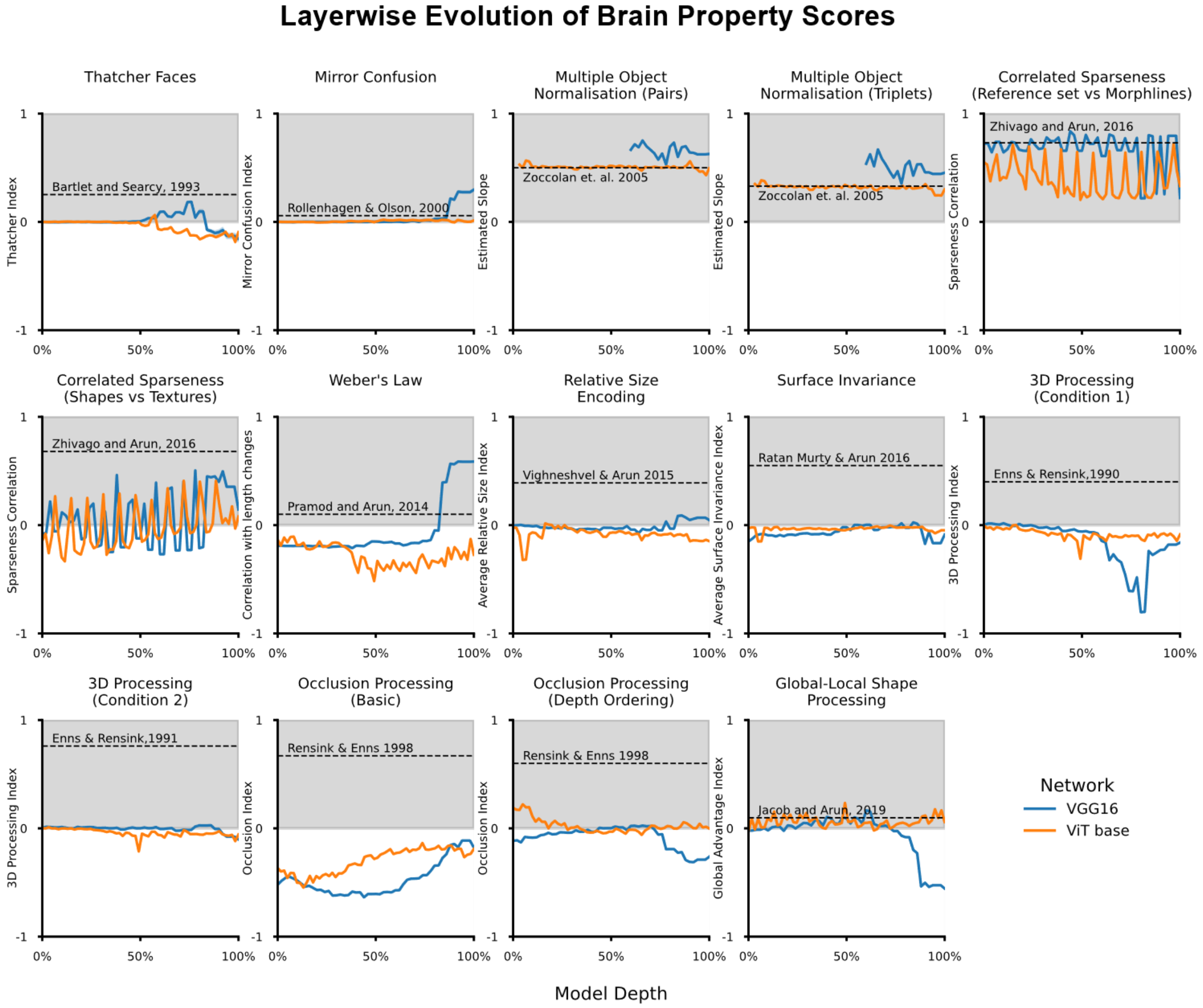

**Figure S2. Layerwise evolution of brain properties.** The layerwise evolution of each tested brain property for VGG16 and ViT-base is shown here. Each subplot corresponds to a brain property. The model depth for each DNN is shown as fractional model depth, calculated as the model layer number divided by the total number of layers for the deep network.

## SECTION S3. RELATION WITH OTHER PERFORMANCE METRICS

We wondered whether the brain property scores of each network would be correlated with its adversarial vulnerability and out-of-distribution (OOD) vulnerability. If any brain property scores are predictive of well-performing networks on these terms, they should be negatively correlated with the metrics.

Adversarial vulnerability was computed by generating adversarial samples using the FGSM method (Goodfellow et al., 2014) and computing the DNN's accuracies on that set $(acc_{adv})$ and comparing it with the accuracy on the original set $(acc_{clean})$. The images used are from the ImageNet validation set and hence, only ImageNet-trained models are used in this analysis. More formally, the adversarial vulnerability metric is as follows:

$$v_{adv} = \frac{acc_{clean} - acc_{adv}}{acc_{clean}}$$

The OOD vulnerability metric was also calculated in the same way by evaluating the DNNs' performance on Imagenet3DCC (Kar et al., 2022). The performance on each corruption set was averaged to get $(acc_{ood})$.

$$v_{ood} = \frac{acc_{clean} - acc_{ood}}{acc_{clean}}$$

Firstly, we find that these two metrics are correlated (Figure S3A). Next, on calculating the correlation between each metric across networks with their respective brain property scores, we found a spectrum of positive and negative correlations, suggesting that the brain property scores capture these metrics.

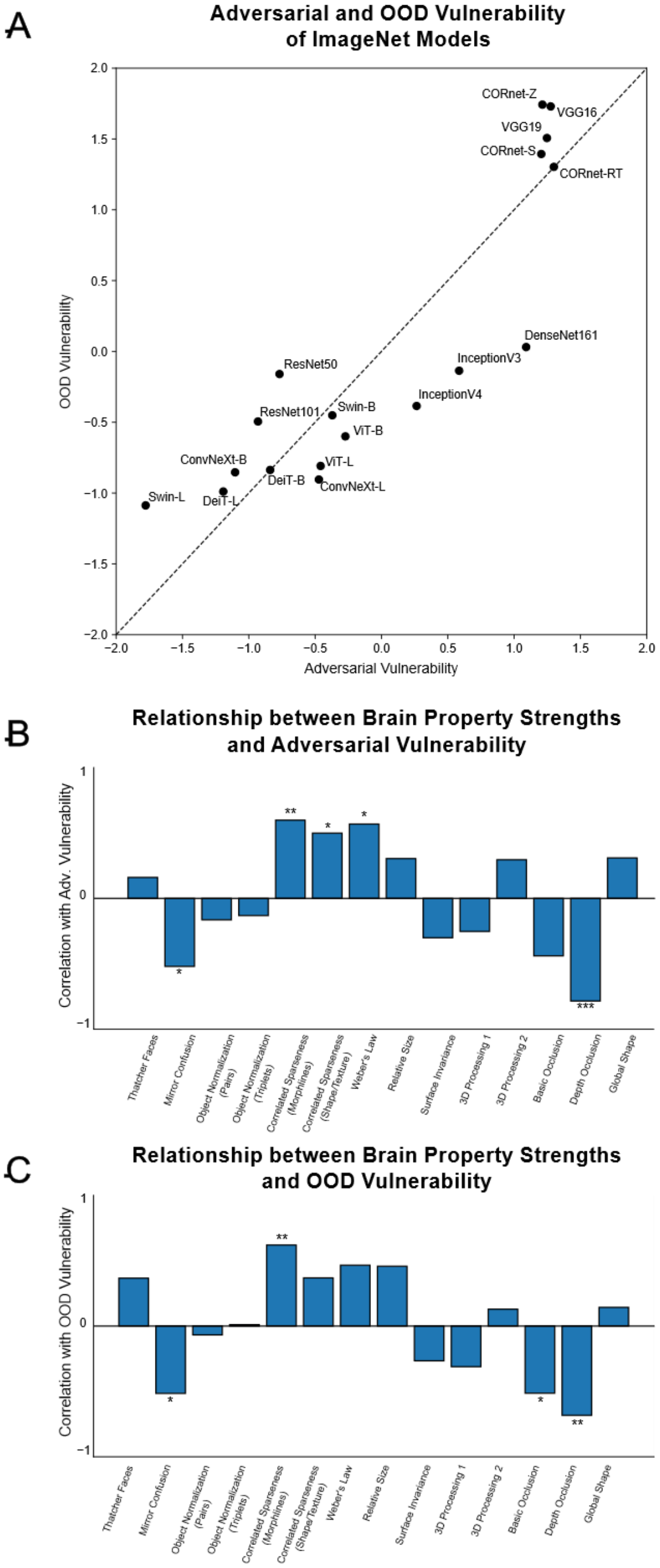

**Figure S3 Relation between brain properties and other performance metrics.**

(A) Correlation between adversarial and OOD vulnerability for the 18 ImageNet DNNs.

(B) Correlation between each brain property score and adversarial vulnerability scores across all models.

(C) Correlations between each brain property score and OOD vulnerability scores across all models.